# HalluSearch at SemEval-2025 Task 3: A Search-Enhanced RAG Pipeline for Hallucination Detection

**Mohamed A. Abdallah*◆, Samhaa R. El-Beltagy*♠**
*School of Information Technology, Newgiza University
◆mohamedabdallah9850@gmail.com
♠samhaa@computer.org

## Abstract

In this paper, we present **HalluSearch**, a multilingual pipeline designed to detect fabricated text spans in Large Language Model (LLM) outputs. Developed as part of Mu-SHROOM, the Multilingual Shared-task on Hallucinations and Related Observable Overgeneration Mistakes, HalluSearch couples retrieval-augmented verification with fine-grained factual splitting to identify and localize hallucinations in 14 different languages. Empirical evaluations show that HalluSearch performs competitively, placing fourth in both English (within the top 10%) and Czech. While the system's retrieval-based strategy generally proves robust, it faces challenges in languages with limited online coverage, underscoring the need for further research to ensure consistent hallucination detection across diverse linguistic contexts.

## 1 Introduction

Ever since the introduction of the transformer architecture (Vaswani et al., 2017) and more specifically with the rise of decoder-only large language models (LLMs), significant advances in the field of natural language processing have been made LLMs excel at text generation and are widely used for tasks such as translation, summarization, and question answering.

Despite their impressive capabilities, being statistical language models, LLMs can sometimes produce factually incorrect or inaccurate statements, often presented in a very convincing manner. This phenomenon is commonly referred to as hallucination. Hallucination is a significant drawback of LLMs which vary in scale from a few billion to hundreds of billions of parameters (Brown et al., 2020). It has been observed that relatively smaller models tend to hallucinate more frequently due to limitations in their training data and model complexity (Li et al., 2024).

Hallucinations in LLMs can have grave consequences. For example, in critical domains like healthcare, relying on an LLM that hallucinates, for diagnosis, can lead to deaths or disabilities. Similarly, in the business and technology sectors, hallucinations may result in poor decision-making, leading to significant financial losses and misallocated investments. Therefore, detecting and localizing hallucinated segments in LLM responses is crucial for developing trustworthy LLM-driven applications.

Mu-SHROOM, is a Multilingual Shared-task on Hallucinations and Related Observable Overgeneration Mistakes (Vázquez et al., 2025) and is and extension to an earlier task, SHROOM (Mickus et al., 2024) and is part of SemEval-2025. The task aims to identify spans of text corresponding to hallucinations within a multilingual context, covering 14 languages: Arabic (Modern standard), Basque, Catalan, Chinese (Mandarin), Czech, English, Farsi, Finnish, French, German, Hindi, Italian, Spanish, and Swedish.

In this shared task, each datapoint consists of an output string generated by one of the publicly available LLMs in response to a user query. The goal is to calculate, on a character-level, the probability that a character is part of a hallucination span.

In this work, we propose a novel system, HalluSearch**,** designed to detect hallucinated spans in multilingual LLM responses. Our approach utilizes retrieval-augmented generation (RAG) as introduced by (Lewis et al., 2020) where external, trusted sources are used to factcheck a model's response which in turn helps determine which parts of a model's response are fabricated, or factual. Our experiments show that HalluSearch achieves competitive performance in several languages such as, English and Czech.

We believe that this is largely due to the availability of online resources related to topics covered in the task. However, challenges persist in consistently detecting hallucinations across other languages, highlighting the need for further refinement and adaptation of the presented approach.

All code and experiment details are publicly available in our GitHub repository, to ensure transparency and reproducibility.

## 2 Related work

As stated earlier, the term 'Hallucination' has been recently used to describe a challenging phenomenon in the context of LLMs. This phenomenon involves generated content that is nonsensical or unfaithful to the input or context provided. The work of (Berberette et al., 2024) re-examines the notion of hallucinations in LLMs through the lens of human psychology. The authors argue that traditional use of this term may be misleading when applied to AI-generated content and emphasize the value of a psychologically informed approach depending on cognitive dissonance, suggestibility, and confabulation as the basis for their approach to mitigate hallucinations and other issues in LLMs.

(Xu et al., 2025) formally addressed hallucination for LLMs by employing results from learning theory findings and demonstrated that hallucination is inevitable for all computable LLMs. HaluEval proposed by (Li et al., 2023) provided a large-scale benchmark for evaluating an LLM's ability to recognize hallucinations. Tonmoy et al. (Tonmoy et al., 2024) surveyed various strategies for mitigating hallucinations, offering insights into potential solutions for this persistent issue.

SHROOM was launched to address the challenge of detecting hallucinations in output generated by Natural Language Generation (NLG) systems. Solutions such as Halu-NLP (Mehta et al., 2024) and OPDAI (Chen et al., 2024) have been proposed to tackle these challenges. However, pinpointing the exact locations of hallucinations was not addressed by this task. Mu-SHROOM, the current iteration of SHROOM, has thus introduced this goal while narrowing the focus to question answering and corresponding LLMs outputs.

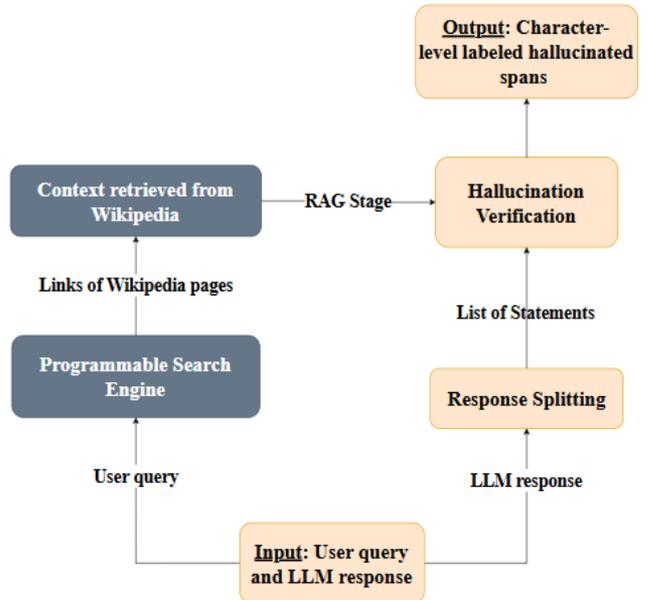

Figure 1: Illustration of HalluSearch building blocks

## 3 Methodology and Experimental setup

Our HalluSearch pipeline includes three major parts: (1) Factual Splitting, (2) Context Retrieval, and (3) Hallucination Verification. The pipeline begins with an input of user query and an LLM-generated response. It then employs the **Factual Splitting** module to divide the text into discrete factual **statements**. Next, the **Context Retrieval** stage uses search results from Wikipedia to gather relevant background information for each query. Finally, the **Hallucination Verification** step compares each statement to its retrieved context counterpart and annotates any spans that looks unsupported or incorrect at a character-level granularity, resulting in an output of labeled hallucinated segments, as shown in figure 1.

### 3.1 Factual Splitting

In this work, we have aimed to break down a LLM response to a query, into segments that ideally each contain a single verifiable proposition. By doing this, our system can isolate discrete claims or statements of a larger LLM-generated response and verify them independently. We have named this step fact splitting.

This approach draws inspiration from earlier work on claim extraction in fact-checking tasks, such as the FEVER dataset (Thorne et al., 2018), where individual claims are identified and evaluated against a knowledge base. Similarly, the notion of **"atomic content units"** introduced in summarization literature (Narayan et al., 2018; Maynez et al., 2020) has shown that decomposing a complex text into smaller, independent factual assertions can greatly improve downstream verification

By applying a similar splitting method in our system, we reduce the risk of conflating multiple assertions within a single verification step minimizing the likelihood of mistakenly flagging correct statements (false positives) or overlooking genuinely incorrect content leaving it undetected within a more extensive chunk of text (false negatives).

The implemented factual splitting module leverages a prompt-based approach to generate atomic claims. Specifically, we employed, the GPT-4o model was used to generate a structured JSON output containing atomic claims and their exact substrings. To achieve this, we crafted a carefully designed prompt that emphasizes the importance of capturing short words and phrases carrying standalone claims (e.g., "Yes," "No," or their multilingual equivalents), while preserving punctuation, spacing, and capitalization.

If a fragment can stand on its own as a separate claim, it is separated from longer statements to facilitate more precise downstream verification. The prompt also enforces strict JSON formatting, which helps us map each extracted statement back to its location in the original text without losing track of language-specific nuances.

Through this fine-grained segmentation, the subsequent verification steps are better positioned to align each claim with authoritative context and detect potential hallucinations on a more granular level.

### 3.2 Context Retrieval

Retrieval-Augmented Generation (RAG) is a powerful technique that enriches language model outputs with external evidence, thereby enhancing factual accuracy and reducing hallucinations (Lewis et al., 2020; Izacard et al., 2021). Rather than relying solely on knowledge learned by LLMs during pretraining, a RAG-based system queries an external knowledge source, such as a search index or document database, to ground its inferences.

This approach has proven to be effective for tasks like open-domain question answering, where up-to-date or domain-specific context is essential. In our pipeline, we apply a similar principle by integrating a Google Custom Search[1] component. This component retrieves potentially relevant web content related to the original input query, allowing our verification model to check each factual statement against live, authoritative sources.

We select the highest-ranked retrieved result for a reputable knowledge source such as Wikipedia, as we aim to obtain the most relevant background information possible. Wikipedia was chosen as a primary knowledge source whenever it appears in the results, due to its status as the most diverse, popular, and widely used encyclopedia globally. Fallbacks are utilized to address incomplete or unavailable search results.

Specifically, two fallback strategies were implemented to retrieve some form of context. In the first strategy, keyword extraction on the user query or statement is carried out to reissue a more concise, focused query to the Custom Search API. This step can be critical in languages or domains where standard queries are too broad, or if initial results are sparse.

If this strategy fails to produce a usable context, we resort to a language model fallback, prompting an LLM (GPT 4o in our case) to generate a short textual passage in the same language as the query.

Although this third option is less reliable in terms of factual grounding, since the LLM itself can hallucinate, it ensures that the verification stage has at least some reference text to work with, Table 1 shows an example of fallback scenarios. Hence, our retrieval architecture remains robust across scenarios where conventional search engines might lack indexed pages or face query limitations, thereby maintaining a RAG-inspired approach in all but the most complicated cases.

### 3.3 Hallucination Verification

The goal of this step is to determine whether each independent claim is a hallucination or not. To carry out this step, each independent claim is paired with a context as detailed in the previous step.

Once each statement is paired with contextual information, the next challenge is fact verification (Thorne et al., 2018; Augenstein et al., 2019). Earlier verification systems relied on specialized classifiers or natural language inference (NLI) models to judge correctness. However, our approach uses a RAG based approach, where an LLM (GPT 4o in our case) is prompted to cross-check each statement against the retrieved context and identify any specific substrings that contradict the context. This is consistent with the approach presented in (Zheng et al., 2024; Wang et al., 2023).

---

[1] https://developers.google.com/custom-search/v1/overview

| |
|---|
| *Example: User Query* |
| Comment a été initialement été appelée la vile de Kaspiisk à sa création? |
| *System Log* |
| No items in Google Search results. |
| Retrying search with extracted keywords: 'vile Kaspiisk création' |
| No items in Google Search results. |
| No context found from Google. Calling LLM to answer the query in the same language. |

Table 1: Handling fallback scenarios with keyword extraction and LLM call.

In this way, our system provides minimal conflicting spans instead of binary labels alone. This allows for more human-interpretable rationales and delivers the incorrect portions of statements in a clearer manner.

The prompt that is passed to the LLM (GPT 4o) is carefully structured system prompt that includes both the source context and a JSON array of factual statements. Detailed guidelines for extracting contradictory substrings help the system handles diverse errors, such as uncertainties or logical inconsistencies, while verifying each factual statement against its retrieved context. In postprocessing, flagged substrings are mapped back to their exact positions in the original text, enabling precise error analysis as detailed in the next subsection.

### 3.4 Postprocessing

After hallucination verification, the flagged substrings must be accurately realigned to the original text. Our postprocessing module achieves this by searching for each extracted substring in the full model output and noting its start and end character indices. We provide two output variants, each aligned with the official Mu-SHROOM metrics. First, a hard-label extractor returns discrete spans for computing intersection-over-union (IoU) against gold annotations.

Second, a soft-label annotator assigns a probability of hallucination. This annotation is consistent with how the data was originally annotated; each span was given a probability of being hallucinated according to the votes it was assigned by the annotators. Soft labels are required to compute the correlation with annotator probabilities. With a single LLM taking the decision, annotation is done by labeling detected hallucinated characters with ones and all others with zeros ensuring compatibility with Mu-SHROOM's rigorous benchmarks.

### 3.5 Variants and Practical Challenges

In addition to mainly depending on GPT-4o[2] closed source model in our experiments, we conducted experiments with open-source models in a voting-style ensemble approach, allowing multiple models to collaboratively vote on detecting hallucination spans in each response as originally done by the annotators, however, results with this approach were not very impressive. We also tested deepseek-reasoner[3] model exclusively for Arabic queries, which achieved better performance on Arabic content than GPT-4o.

Our goal was to establish a generic system that is robust to multilingual data. However, implementation challenges arose when dealing with 14 languages, each featuring distinct morphological rules and varying levels of web coverage. Certain languages, like Basque or Farsi, have limited online resources which reflected adversely on search engine results. This limitation forced reliance on fallback strategies such as LLM-based context generation risking further hallucination.

Moreover, when extracting keywords from morphologically rich, under-resourced languages nonsensical or misleading outputs hinder effective context retrieval. These factors can undermine retrieval success and make robust coverage more difficult to achieve, degrading the system performance. This highlights the complexity of the multilingual hallucination detection task, pushing for more robust fallback strategies and meticulous pre and post processing steps.

---

[2] https://platform.openai.com/docs/models/gpt-4o
[3] https://api-docs.deepseek.com/guides/reasoning_model

| Language | IOU | Cor | Rank |
|---|---|---|---|
| AR | 0.5362 | 0.5258 | 10 |
| CA | 0.5215 | 0.5704 | 11 |
| CS | 0.4911 | 0.4942 | 4 |
| DE | 0.5187 | 0.5056 | 13 |
| EN | 0.5656 | 0.5360 | 4 |
| ES | 0.3883 | 0.4456 | 12 |
| EU | 0.5251 | 0.4789 | 6 |
| FA | 0.4443 | 0.4734 | 14 |
| FI | 0.5681 | 0.5297 | 12 |
| FR | 0.4366 | 0.3365 | 20 |
| HI | 0.5265 | 0.5195 | 13 |
| IT | 0.5484 | 0.5604 | 14 |
| SV | 0.5622 | 0.4290 | 8 |
| ZH | 0.4534 | 0.4232 | 13 |

Table 2: Performance metrics over test data across multiple languages

## 4 Results

In our experiments (Table 2), HalluSearch exhibits strong performance in several languages, notably ranking **4th** on **English** and **Czech** (complete results are found in Mu-SHROOM's original paper, Vázquez et al., 2025). We observe that prompt refinements, such as adding chain-of-thought reasoning instructions, can yield significant gains, with GPT-4o improving English results. Using 'deepseek-reasoner' model boosts Arabic performance.

Conversely, attempts to combine multiple open-source models and voting ensemble did not enhance results, possibly due to inconsistent alignment between the models' annotations. Overall, HalluSearch's approach to fact verification demonstrates competitiveness across diverse languages, but the variation in rank shows the complexities that remain an open challenge in multilingual hallucination span detection.

## 5 Conclusion

In this work, we have presented HalluSearch, the aim of which was to address the problem of detecting LLM hallucinations. HalluSearch is a search-enhanced RAG pipeline that pinpoints potentially fabricated or incorrect spans in multilingual outputs. By using precise factual splitting, context retrieval from reliable sources, and a prompt-based verification step, our system provides both hard-label and soft-label annotations for hallucination spans. Evaluation results demonstrate that HalluSearch competes well in multilingual settings despite inherent difficulties such as limited content availability in less-resourced languages.

These findings highlight the importance of robust, cross-lingual retrieval strategies and careful prompt engineering. Future research will delve into addressing low-resource languages more effectively, improving fallback mechanisms, and exploring fine-grained alignment techniques for enhanced span detection accuracy.